\documentclass{article}

\usepackage[preprint]{corl_2026} %

\title{Imitation from Heterogeneous Demonstrations using Grounded Latent-Action World Models}

\author{
  Tianyou Wang \quad Anson Lei \quad Joe Watson \quad Ingmar Posner \\
  University of Oxford \\
  \texttt{\{tianyou, anson, joewatson, ingmar\}@robots.ox.ac.uk}
}
\usepackage{hyperref}
\hypersetup{
  colorlinks=true,
  linkcolor=blue,
  citecolor=blue,
  urlcolor=blue,
  filecolor=blue,
}
\usepackage{enumitem}
\usepackage{amsmath}
\usepackage{amssymb}
\usepackage{cleveref}

\usepackage{amsmath,amsfonts,bm}

\def\eqref#1{equation~\ref{#1}}

\def\1{\bm{1}}

\def\vzero{{\bm{0}}}

\def\vtheta{{\bm{\theta}}}
\def\vomega{{\bm{\omega}}}
\def\vphi{{\bm{\phi}}}
\def\vpsi{{\bm{\psi}}}
\def\veta{{\bm{\eta}}}
\def\va{{\bm{a}}}

\def\vc{{\bm{c}}}

\def\vh{{\bm{h}}}

\def\vo{{\bm{o}}}

\def\vs{{\bm{s}}}

\def\vx{{\bm{x}}}
\def\vy{{\bm{y}}}
\def\vz{{\bm{z}}}

\def\mA{{\bm{A}}}

\def\mI{{\bm{I}}}

\def\mO{{\bm{O}}}

\def\mS{{\bm{S}}}

\def\mX{{\bm{X}}}

\def\mZ{{\bm{Z}}}

\DeclareMathAlphabet{\mathsfit}{\encodingdefault}{\sfdefault}{m}{sl}
\SetMathAlphabet{\mathsfit}{bold}{\encodingdefault}{\sfdefault}{bx}{n}

\def\gC{{\mathcal{C}}}
\def\gD{{\mathcal{D}}}

\def\gJ{{\mathcal{J}}}

\def\gL{{\mathcal{L}}}

\def\gN{{\mathcal{N}}}

\def\sD{{\mathbb{D}}}

\def\sR{{\mathbb{R}}}

\newcommand{\E}{\mathbb{E}}

\newcommand{\KL}{\sD_{\mathrm{KL}}}

\usepackage{graphicx}
\usepackage{booktabs}
\usepackage{wrapfig}
\begin{document}
\maketitle

\begin{abstract}
    Imitation learning has emerged as a powerful paradigm for learning visuomotor policies, but its generalisation and stability are limited by the scale and quality of demonstration data needed.
    A promising direction is to leverage more abundant but heterogeneous data sources, which differ in action space and often lack action labels altogether.
    Existing co-training approaches that combine heterogeneous data sources rely on heuristic and hand-engineered alignment techniques.
    In contrast, we argue that action representations should be grounded in prediction:
    actions that produce the same effect on the environment should share the same representation, regardless of their sources.
    To this end, we instantiate this principle by using a grounded latent-action world model ({GLAM}), a pair of generative models with a shared latent action space across data sources that is grounded by predicting future observations consistently across sources.
    This latent action space is used to train downstream behavioural cloning (BC) policies which map observations to latent actions and decode them back to robot actions, providing a paradigm for learning from heterogeneous data.
    Empirically, we demonstrate that GLAM successfully learns an aligned latent action space that facilitates action transfer across data sources with and without action labels.
    Across five manipulation tasks in simulation and in the real world, GLAM-aligned policies significantly outperform BC baselines and prior latent-action methods, achieving an average of +48\% improvement in task success rate with the same data-scarce setting. Videos and code are available at \url{https://viccccciv.github.io/glam/}.
\end{abstract}


\section{Introduction}

Modern imitation learning has produced impressive visuomotor manipulation policies~\citep{diffusion,act,mip,pi0}, yet their generalisation remains tightly coupled to the scale and quality of expensive demonstration data. Even on simple tabletop tasks, behaviour cloning typically requires hundreds to thousands of teleoperated trajectories on the target robot to cross the threshold of reliable deployment~\citep{lin2025data,equivariant,mimicgen}. Collecting such data is slow and costly, and the dominant bottleneck for scaling capable policies to the long tail of real-world tasks. One promising approach to sidestep this problem is to replace some of this expensive supervision with cheaper data, such as trajectories collected with portable hand-held devices~\citep{umi,dexcap,dexumi}, in simulation~\citep{simrealct,mimicgen}, or scraped from human video~\citep{r3m,zeromimic}. These sources are abundant and far cheaper than demonstrations on the target robot. The central aim of this paper is to investigate how one can leverage heterogeneous data to supplement target-robot demonstrations.

Using heterogeneous data is challenging: different data sources carry different action spaces, frequently lack action labels altogether, and exhibit substantial visual variation in embodiment and scene that does not reflect the underlying task. These mismatches mean that naively pooling the data and training on the mixture rarely works. Prior work bridges these gaps with hand-crafted alignment recipes~\citep{ot_sim,immimic,egobridge,humanoid}. For example, the sim-real co-training method~\citep{ot_sim} aligns simulation and real feature spaces with an optimal-transport loss, a DTW-based temporally-aligned sampling strategy, and an empirically tuned mixing ratio. In contrast, we advocate for a more principled approach: we posit that \emph{what matters for manipulation is how an action affects the environment, regardless of where it comes from}. Two actions that drive the manipulated object along the same trajectory should be represented similarly, regardless of whether they originate from different embodiments, real or simulated. This turns cross-source integration into a representation-learning problem, with environment transitions as the shared, physically grounded supervisory signal.

We argue that a world model, trained to predict how actions shape environment transitions, provides a natural way to ground latent actions in physical dynamics. Given labelled demonstrations in the target domain and auxiliary demonstrations without action labels, we treat actions as latent variables and learn two coupled generative models (grounded latent-action world model, GLAM): one over the full mixture of target and auxiliary data, and one over the labelled target data. The mixed-data model uses an inverse dynamics model (IDM) to infer latent actions directly from environment transitions, enabling source-invariant inference without action labels. The target-domain model uses an action encoder to infer latent actions from executable robot actions, ensuring that the latent space remains grounded in the target robot’s action space. An asymmetric KL objective, together with shared forward dynamics, binds these two inference pathways into a single aligned latent action space.
We then propose a GLAM-aligned behavioural cloning (BC) pipeline based on this unified action space. Specifically, using GLAM, we relabel all available data with source-agnostic, control-aware latent actions that serve as supervision signals for a downstream BC policy. 

Empirically, we validate that GLAM unifies labelled and unlabelled actions into a single latent space across data sources. Latent actions inferred from auxiliary demonstrations can be transferred directly into executable actions for a target robot arm, making unlabelled auxiliary data usable as BC supervision. Moreover, we experimentally demonstrate that GLAM-aligned latent action policies are able to efficiently learn from auxiliary data sources, outperforming BC and other latent action baselines (up to $+69\%$) across three real-world and two simulated manipulation tasks. 
\section{Related Work}
\label{sec:related_work}

\paragraph{Imitation Learning from Heterogeneous Demonstrations.}
Modern robotic imitation learning spans action chunking~\citep{act}, diffusion policies~\citep{diffusion}, regression-based variants~\citep{mip}, behaviour transformers~\citep{vqbet}, and generalist VLA models~\citep{openvla,pi0,octo,gr2}, all of which rely on large amounts of action-labelled target-robot demonstrations. To reduce this dependence, recent work explores two directions for leveraging cheaper sources. A first line learns from auxiliary data alone: visual representation pretraining from web video relaxes the perception side~\citep{r3m,vcl,vip}, while portable hand-held devices~\citep{umi,dexcap,dexumi} and in-the-wild human videos~\citep{vidbot,track2act,H2R} supply trajectories collected entirely off the target robot. Hand-held devices, however, need extra hardware-matching to the target gripper and still rely on human collection, which remains time-costly. Human videos drop the device altogether but, lacking target-robot grounding, yield embodiment-agnostic affordances that struggle with precision, dexterity, and task-level learning. A second line co-trains the policy on a mixture of target and auxiliary sources to reduce teleoperation cost while preserving target-specific supervision~\citep{inNon,simrealct,egomimic,openxe,egobridge}. For example, prior work~\citep{simrealct} co-trains a single policy on a mixture of simulation and real data, relying on an empirically tuned mixing ratio and hand-aligned camera viewpoints between simulation and the real-world setup. These approaches typically navigate data mismatches through empirical mixing ratios, source-specific tokenizers, or heuristic and engineered embodiment alignment, treating cross-source integration as a tuning rather than a learning problem. GLAM instead leverages the world model's nature of modeling action and environment interactions to unify heterogeneous data into a shared physics space.

\paragraph{From Imagined Rollouts to Latent-Action Anchoring.}
 World models in robot learning are predominantly used as imagined simulators for policy optimization~\citep{ha2018world,daydreamer,storm} or as planners for trajectory optimization~\citep{lewm,dinowm,tdmpc,vjepa2,pldm}. Both routes depend on substantial online interaction or massive pretraining to make rollouts reliable enough for control. GLAM instead uses the world model as a frozen latent-action anchor that supplies only per-step labels rather than rolled-out trajectories, sidestepping the online interaction needed by RL and the autoregressive drift driving the data demands of trajectory optimization, making the world model trainable from a few hundred mixed trajectories. Closer to our setting, Latent Action Models (LAMs) supply pseudo-action labels for behaviour cloning, extracting action latents from inter-frame transitions. Discrete latents tokenize inter-frame transitions for downstream policy learning~\citep{lapa,villax,vpt,igor,moto,amplify,univla}, while continuous latents directly supervise downstream policies via regression~\citep{lapo,clam,lawm,uniskill,dynamo}. Two representatives sharpen the design choice: LAPA~\citep{lapa} learns a discrete VQ codebook and utilizes it via large-scale VLM finetuning; CLAM~\citep{clam} uses a continuous latent jointly trained with an action decoder for direct grounding, but in the single-embodiment setting. GLAM differs in three ways: (i) a single IDM trained across both target and auxiliary data delivers a source-invariant latent action space, rather than one tied to a single embodiment; (ii) a target-side action encoder bound to the IDM by asymmetric KL alignment and shared forward dynamics injects control semantics into the latent, avoiding post-hoc grounding of pixel-reconstruction latents loosely tied to actions; (iii) the resulting model serves as a frozen latent-action anchor that turns unlabelled auxiliary data into BC supervision, without large-scale pretraining.

\section{Learning from Heterogeneous Demonstrations with World Models}
\label{sec:method}
\begin{figure}[t]
    \centering
    \includegraphics[width=\linewidth]{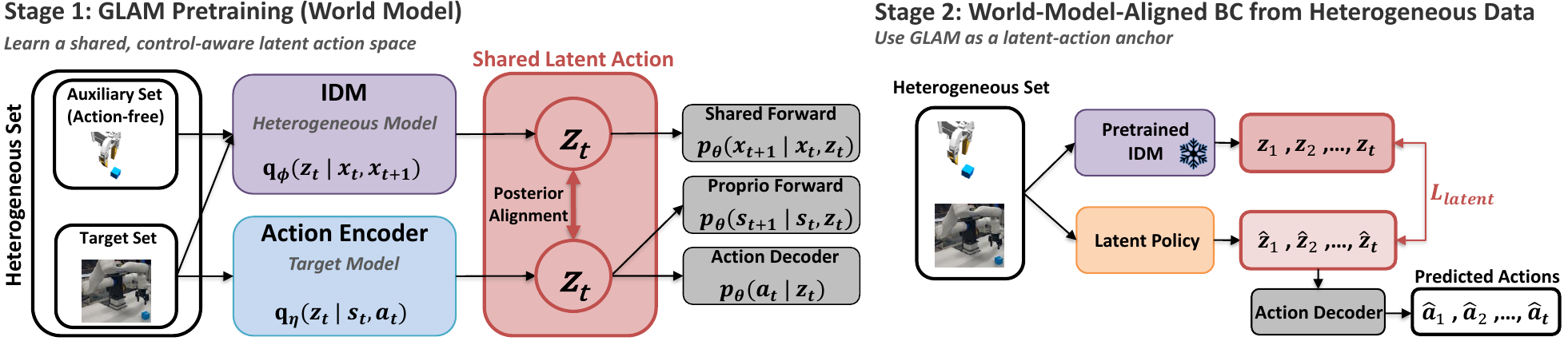}
    \caption{\textbf{GLAM-aligned imitation learning pipeline.}
    \textbf{Stage 1 (left):} GLAM is pretrained on a heterogeneous demonstration set; an IDM (posterior) and an action encoder (posterior) over a shared latent action $\vz_t$ are aligned by an asymmetric KL and grounded by shared forward dynamics. 
    \textbf{Stage 2 (right):} The frozen GLAM relabels every transition with $\vz_t$, which supervises a downstream BC policy that predicts latent action chunks and decodes them into executable actions.}
    \label{fig:pipeline}
\end{figure}

We aim to learn a visuomotor policy on a target robot that benefits from auxiliary data $\mathcal{D}^{\mathrm{aux}}$ in addition to a small set of target-robot data $\mathcal{D}^{\mathrm{tar}}$.
We assume access to the heterogeneous dataset $\mathcal{D} = \mathcal{D}^{\mathrm{tar}} \cup \mathcal{D}^{\mathrm{aux}}$. $\mathcal{D}^{\mathrm{tar}}$ contains trajectories of  $\tau = \{(\vo_t, \vs_t, \va_t)\}_{t=1}^{T}$, where $\vo_t$ is the observation including the image and end-effector pose, $\vs_t$ the proprioceptive robot state, and $\va_t$ the robot action.  
Note that $\mathcal{D}^{\mathrm{aux}}$ lacks action labels $\va_t$ and robot state $\vs_t$, so it contains trajectories of $\tau = \{\vo_t\}_{t=1}^{T}$.
We let auxiliary data co-supervise the downstream BC policy through a two-stage pipeline (\Cref{fig:pipeline}). In Stage 1 (\Cref{sec:hrwm}), we treat actions as a latent variable and formulate a pair of generative models (GLAM): one for the heterogeneous dataset $\mathcal{D}$, and the other for the target dataset $\mathcal{D}^{\mathrm{tar}}$, grounding the latent actions inferred from each model into a shared space. In Stage 2 (\Cref{sec:policy}), we relabel every transition in $\mathcal{D}$ with its latent action and train a BC policy to predict latent actions, decoded back to executable robot actions via an action decoder. 

\subsection{Grounded Latent-Action World Models}
\label{sec:hrwm}

\paragraph{A generative-model view of latent actions.}
Stage 1 (\Cref{fig:pipeline}, left) instantiates the shared latent action as a continuous variable $\vz_t \in \sR^d$ in two generative models over trajectory distributions:
\begin{align}
\label{eq:gen-h} \text{(heterogeneous)}\;\;
p(\mX, \mO, \mZ) &= p(\vx_1)\textstyle\prod_{t=1}^T p(\vx_{t+1} | \vx_t, \vz_t)\,p(\vo_t|\vx_t)\,p(\vz_t), \\
\label{eq:gen-t} \text{(target)}\;\;
p(\mS, \mA, \mZ) &= p(\vs_1)\textstyle\prod_{t=1}^T p(\vs_{t+1} | \vs_t, \vz_t)\,p(\va_t | \vz_t)\,p(\vz_t),
\end{align}
where upper-case random variables refer to trajectories, e.g., $\mO = [\vo_1, \dots, \vo_T]$, and $\vx_t$ is the latent state encoded from observations.
The heterogeneous generative model has access only to observations $\mO$ and is structured as a standard latent space dynamics model~\citep{rssm} with a latent action variable. 
Importantly, we use both $\gD^{\mathrm{tar}}$ and $\gD^{\mathrm{aux}}$ to jointly train this single generative model, ensuring that the transitions in both datasets are generated by latent action $\vz_t$ from a unified space.
To train this heterogeneous model, we instantiate two learnable posteriors over the latent state and latent actions, ${q_\vpsi(\vx_t\mid\vo_t)}$ and ${q_\vphi(\vz_t\mid\vx_t,\vx_{t+1})}$, where the posterior over latent state $\vx_t$ is a standard image encoder and the posterior over latent action $\vz_t$ is an \textit{inverse dynamics model} (IDM) that infers the action variable given the state transition. Crucially, the use of a shared inverse dynamics model and a shared forward model $p_\vtheta(\vx_{t+1} | \vx_t, \vz_t)$ across the target and the auxiliary datasets ensures source-invariance of the latent action space, implementing our core principle that any action, regardless of the dataset of origin, should be represented based on how it affects the environment transition.
This first generative model allows us to optimise the evidence lower bound objective (ELBO) \citep{vae},
\begin{align}
\label{eq:elbo-h}
\gJ_{\mathrm{H}}(\vphi, \vpsi, \vtheta, \gD)
= \E\,\big[&\textstyle\sum_{t=1}^T 
   \E\left[\log p_\vpsi(\vo_{t} | \vx_t) - \KL\left(q_\vpsi(\vx_{t+1} |\vo_{t+1}) \,\|\, p_\vtheta(\vx_{t+1} | \vx_t, \vz_t)\right)
   \right] \notag\\ &\hspace{2em}- \KL\big[q_\vphi(\vz_t | \vx_t, \vx_{t+1}) \,\|\, p(\vz_t)\big]
   -\KL[q_\vpsi(\vx_1|\vo_1)\mid\mid p(\vx_1)]
\big],
\end{align}
where observation trajectory $\mO$ is sampled from $\gD$ and $\vz_t, \vx_t, \vx_{t+1} \sim 
q_{\vphi,\vpsi}(\cdot|\vo_t, \vo_{t+1})$
where 
$q_{\vphi,\vpsi}(\vz_t, \vx_t, \vx_{t+1}|\vo_t, \vo_{t+1}) = 
q_\vphi(\vz_t| \vx_t, \vx_{t+1})\, q_\vpsi(\vx_t|\vo_t)\,q_\vpsi(\vx_{t+1}|\vo_{t+1})$. This formulation is similar to the Dreamer world model architecture~\citep{dreamerv3}, with an additional IDM posterior that infers the latent action from observed transitions.

The second generative model grounds its latent action space using extra supervision signals from the robot state $\mS$ and action labels $\mA$, available only on $\gD^{\mathrm{tar}}$.
This ELBO jointly learns the \textit{action encoder} $q_\veta(\vz_t\mid\vs_t,\va_t)$ and dynamics model $p_\vtheta(\vs_{t+1}\mid\vs_t,\vz_t)$,
{\setlength{\jot}{0.5pt}
\begin{align}
\label{eq:elbo-t}
\gJ_{\mathrm{T}}(\vtheta, \veta, \gD^{\mathrm{tar}})
= \E\,\big[&\textstyle\sum_{t=1}^T 
   \E\left[\log p_\vtheta(\vs_{t+1} | \vs_t, \vz_t) + \log p_\vtheta(\va_t | \vz_t)\right]\notag\\
   &\hspace{14em}- \KL\big[q_\veta(\vz_t | \vs_t, \va_t) \,\|\, p(\vz_t)\big]
\big],
\end{align}}

where $(\mS, \mA)$ is sampled from $\gD^{\mathrm{tar}}$ and $\vz_t \sim q_\veta(\cdot | \vs_t, \va_t)$. The action encoder posterior together with the target generative model injects control-awareness into the action latent, ensuring it contains the information needed to recover the executable robot action in the target domain. The second action-reconstruction term $\log p_\vtheta(\va_t \mid \vz_t)$ enforces this latent-to-action decodability and is attached to the action encoder branch only, so that the IDM in \Cref{eq:elbo-h} remains free of target-specific control particularities. In our implementation, we set the prior, $p(\vz_t)$, to $\gN(\vzero, \mI)$ for both generative models. Implementation details of the exact model architecture are included in \Cref{app:impl}.

\paragraph{Asymmetric alignment between posteriors.}
Finally, in order to bridge the pair of generative models and their respective approximate posteriors, we introduce an additional soft alignment constraint, $\gC_{\text{KL}}$, that encourages agreement between the two posteriors over the target dataset,
\begin{equation}
\label{eq:align}
\gC_{\mathrm{KL}}(\vphi, \veta, \vpsi, \gD^{\mathrm{tar}}) \;=\; \E_{\mX\sim q_\vpsi(\cdot|\mO),\,\mO, \mS, \mA \sim \gD^{\mathrm{tar}}} \Big[\KL \big[q_\vphi(\vz_t | \vx_t, \vx_{t+1}) \,\|\, \mathrm{sg}[q_\veta(\vz_t | \vs_t, \va_t)]\big]\Big],
\end{equation}
where $\mathrm{sg}[\cdot]$ denotes the stop-gradient operation. This deliberate asymmetry reflects that the action encoder carries privileged target-only signal, namely, the ground-truth action labels in the target space.
Through this alignment, the IDM absorbs executable-action semantics on target transitions and transports them back to auxiliary ones. This preserves the IDM's source-invariance while keeping the action encoder undistorted by the more ambiguous IDM signal.

\paragraph{Training objective.}
In summary, GLAM is trained end-to-end by jointly minimising the two negative ELBO losses with respect to the latent action encoders, state encoders and dynamics models, 
\begin{align}
\gL(\vphi,\veta,\vpsi,\vtheta,\gD, \gD^{\mathrm{tar}}) \;=
-\gJ_{\mathrm{H}}(\vphi,\vpsi,\vtheta,\gD) -
\gJ_{\mathrm{T}}(\vtheta,\veta,\gD^\text{tar}) + \lambda\,\gC_{\mathrm{KL}}(\vphi,\veta,\vpsi,\gD^{tar}),
\label{eq:total}
\end{align}
where soft constraint weighting $\lambda \geq 0$.
When combined, these objectives unify $\gD^{\mathrm{tar}}$ and $\gD^{\mathrm{aux}}$ into a single, source-invariant, control-aware latent space from which any transition can be relabelled with a shared latent action.

\paragraph{Object masks as observations.}
Our hypothesis is that latent actions should be mapped to the same point if they affect the environment equally.  In manipulation tasks, we can further sharpen this: the action representation should reflect how the manipulated object moves. To test this hypothesis, we also investigate a variant of our model that uses binary segmentation mask of the manipulated object extracted by an off-the-shelf segmentation model~\citep{sam3} in place of the RGB frame without changing any ELBO or the training objectives above. In \Cref{sec:ablation}, we empirically show that object masked transitions lead to further improved transfer between auxiliary and target dataset.
\subsection{World-Model-Aligned Imitation Learning}
\label{sec:policy}
Building on the pretrained GLAM in \Cref{sec:hrwm}, we introduce a world-model-aligned imitation learning pipeline that turns heterogeneous data into BC supervision. The key idea is to learn latent policies that map observations to the GLAM aligned latent actions, and then decode the policy outputs to robot actions for execution. Concretely, we first relabel the available data, both auxiliary and target, using the IDM posterior mean as action labels,
\begin{equation}
\label{eq:label}
\vz_t \;\leftarrow\; \mu_\vphi^{\mathrm{IDM}}(\vx_t, \vx_{t+1}),
\end{equation} 
which provides a large dataset of observation action pairs that serves as supervision signals for a downstream BC policy. Intuitively, the latent action labels distill information about how objects need to be manipulated for the task at hand while discarding source-specific information.

In principle, any downstream BC architectures can be used for learning policies from our augmented dataset. The specific policy structure used in our experiments is illustrated in \Cref{fig:pipeline} (right).
The policy operates directly on raw RGB images at both training and inference time. Specifically, two camera views are encoded by a jointly trained DINOv2 backbone~\citep{dinov2}, followed by a small learned projector, and concatenated with the proprioceptive state $\vs_t$ to form the policy input $\vc_t$. We adopt MIP~\citep{mip}, a lightweight two-step regression policy that combines stochasticity injection with supervised iterative computation, as our latent policy. MIP has been shown to match the performance of generative control policies, e.g.\ diffusion policies~\citep{diffusion}, while requiring no distribution fitting. Our latent-action policy predicts a chunk of latent actions $\hat{\vz}_{t:t+H} \sim \pi_\vomega(\,\cdot \mid \vc_t\,)$, which are further decoded into executable actions as $\hat{\va}_{t:t+H} = \vh_\vomega^a(\hat{\vz}_{t:t+H})$.
The policy is trained end-to-end on $\mathcal{D}^{\mathrm{tar}} \cup \mathcal{D}^{\mathrm{aux}}$ with the objective:
\begin{equation}
\label{eq:policy-loss}
\gL_\pi(\vomega,\gD^{\mathrm{tar}},\gD^{\mathrm{aux}}) \;=\; \E_{\tau \sim \gD}\!\big[ \| \hat{\vz}_{t:t+H} - \vz_{t:t+H} \|_2^2 \big] \;+\; \E_{\tau \sim \gD^{\mathrm{tar}}}\!\big[ \| \hat{\va}_{t:t+H} - \va_{t:t+H} \|_2^2 \big].
\end{equation}
The first term draws supervision from all of $\gD = \gD^{\mathrm{tar}} \cup \gD^{\mathrm{aux}}$, so auxiliary trajectories shape $\pi_\vomega$ on equal footing with target ones; the second term is restricted to $\gD^{\mathrm{tar}}$ where action labels are available.
\section{Experiments}
\label{sec:result}

In this section, we present experimental validation of the proposed pipeline. We aim to answer two questions: first, whether GLAM learns a shared latent action space that is consistent across different data sources; and second, whether auxiliary, non action-labelled data can be used to augment training datasets and improve BC performance. We address the first question qualitatively in \Cref{sec:qual-wm} through investigating cross-source latent action transfer, and the second in \Cref{sec:quant} through comparisons against BC and latent-action baselines.

\begin{figure}[t]
    \centering
    \includegraphics[width=0.85\linewidth]{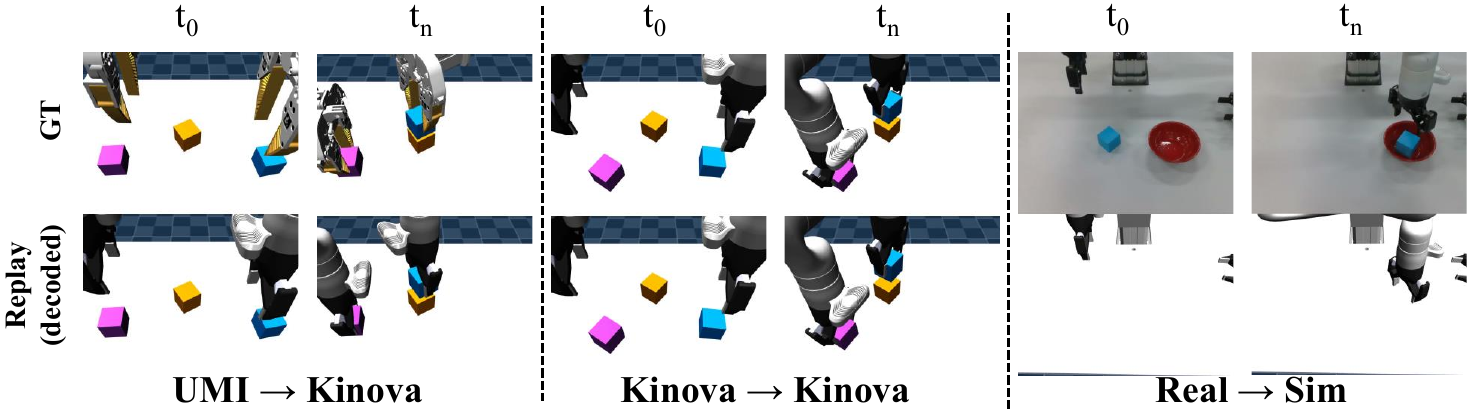}
    \caption{\textbf{Cross-source transfer through the shared latent action space.} For each group, an unseen episode (top: GT) is encoded by the IDM into latent actions, decoded by $p_\vtheta(\va_t \mid \vz_t)$, which has only seen Kinova-sim data, and replayed open-loop on Kinova in simulation (bottom). Latents from \textbf{UMI}, \textbf{Kinova-sim}, and \textbf{Kinova-real} episodes all reproduce the original motion on Kinova-sim, validating action space alignment in cross-embodiment, in-distribution, and sim-real settings.}
\label{fig:cross-source-transfer}
\end{figure}

\subsection{Experimental Setup}
\label{sec:exp-setup}
\textbf{Tasks.} 
We evaluate our pipeline on five manipulation tasks spanning two domains. In the real world, we implement three tasks: \emph{lifting}, picking up a cube; \emph{pick-and-place}, transporting a cube into a bowl; \emph{knock-down}, knocking down a mustard bottle, which requires the gripper to approach at the correct angle since the bottle is stable when struck on its side.
In simulation we use two tasks: \emph{2-cube stacking} with a single arm and \emph{3-cube stacking} with two arms. \\
\textbf{Robot Platforms.} 
The real-robot setup uses a 7-DoF single Kinova arm with a parallel-jaw gripper (Kinova), observed from two calibrated RealSense cameras. Simulation uses MuJoCo, where the single-arm tasks employ the same Kinova model as in the real setup, and the two-arm task employs two such arms in a shared workspace. The auxiliary data source uses a floating UMI gripper~\citep{umi} simulated in MuJoCo without joint actions, which is to test whether GLAM can cross this large domain gap. \Cref{fig:setup} shows the hardware and per-task visualizations.\\
\textbf{Datasets.}
For each task, we collect a small target set  $\mathcal{D}^{\mathrm{tar}}$ of 100 demonstrations and a larger auxiliary set $\mathcal{D}^{\mathrm{aux}}$ of 400 unlabelled trajectories. For the real world tasks, $\mathcal{D}^{\mathrm{tar}}$ consists of 100 teleoperated Kinova trajectories; $\mathcal{D}^{\mathrm{aux}}$ consists of 400 simulated UMI trajectories. The world model and policy are trained on $\mathcal{D}^{\mathrm{tar}} \cup \mathcal{D}^{\mathrm{aux}}$ and evaluated on the real robot. We additionally use 100 Kinova-sim trajectories as a third optional auxiliary set for multi-source experiment (\Cref{tab:third-source}). For simulated tasks, $\mathcal{D}^{\mathrm{tar}}$ comprises 100 simulated Kinova trajectories and $\mathcal{D}^{\mathrm{aux}}$ 400 UMI trajectories, with the world model and policy trained on both datasets and evaluated in simulation.\\
\textbf{Baselines.}
To isolate the contribution of our world model design, we compare against prior latent action works that are implemented with the same network architecture, the same downstream policy, and the same training-data scale as GLAM. Concretely, for each prior work we extract its world-model components (the IDM, the FDM, and the action heads) and the only difference is the latent-action structure learned. The world models are then plugged into the same MIP-based BC policy as ours. Concretely, we evaluate the following models:
\begin{itemize}[leftmargin=*, itemsep=1pt, topsep=0pt, parsep=0pt, partopsep=0pt]
    \item \textbf{MIP~\citep{mip}.} A regression-based behaviour cloning policy. Since MIP requires action labels, we train it on the 100 target-robot trajectories of $\mathcal{D}^{\mathrm{tar}}$ alone.
    \item \textbf{CLAM~\citep{clam}.} A latent action model that learns a continuous latent IDM and FDM together with a jointly trained action decoder. As CLAM is designed for the single-embodiment setting, we train it on $\mathcal{D}^{\mathrm{tar}}$, label transitions with its IDM and train the MIP policy to predict these latents.
    \item \textbf{LAPA~\citep{lapa}.} A VQ-VAE-based latent action model that learns a discrete codebook of inter-frame latent actions through next-frame reconstruction. Because the codebook is shared across all data sources by construction, we train its IDM and FDM on both $\mathcal{D}^{\mathrm{tar}}$ and $\mathcal{D}^{\mathrm{aux}}$, label every transition with its quantized latent, which the downstream MIP is trained to predict.
    \item \textbf{CLAM-O and LAPA-O.} We evaluate two object-masked variants of CLAM and LAPA, in which the world model input is replaced by the same binary object mask used in our method, isolating the contribution of our paired generative models from the object-mask advantage.
    \item \textbf{GLAM and GLAM-O.} GLAM uses our world model (\Cref{sec:hrwm}) with raw RGB input; GLAM-O is the same model with the input replaced by an object-centric binary mask.
\end{itemize}
\vspace{-6pt}

\subsection{Cross-Source Transfer through a Shared Latent Action Space}
\label{sec:qual-wm}
To assess whether GLAM has learned a shared latent action space across sources, we investigate whether latent actions inferred from an unlabelled source can be executed on the target robot through its action decoder $p_\vtheta(\va_t \mid \vz_t)$. For an unseen episode from a given source, we feed each transition into the IDM to obtain $\vz_t$ by \Cref{eq:label}, decode them through $p_\vtheta(\va_t \mid \vz_t)$, and replay the result open-loop on Kinova in simulation. Since $p_\vtheta(\va_t \mid \vz_t)$ is trained only on $\mathcal{D}^{\mathrm{tar}}$, latents from any other source must land within its training distribution to decode coherently. \Cref{fig:cross-source-transfer} qualitatively demonstrates that the IDM maps all three sources (UMI, Kinova-sim, Kinova-real) into a unified latent action space, enabling the transfer of behaviours across embodiments \textit{without} any action labels. \Cref{app:cross-source} shows more cross-source transfer results.

\subsection{World-Model Anchoring Enables BC to Generalize from Heterogeneous Data}
\label{sec:quant}
\begin{figure}[t]
    \centering
    \includegraphics[width=0.9\linewidth]{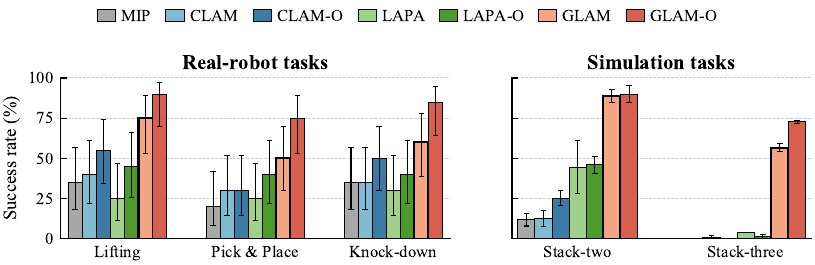}
    \caption{\textbf{Main results across three real-robot and two simulated manipulation tasks.}
    Success rate (\%) of baselines and our method.
    \textbf{Real tasks} are evaluated over 20 trials per task; error bars show 95\% Wilson score intervals~\citep{wilson}.
    \textbf{Simulation tasks} are trained with 3 training seeds and evaluated with 50 trials each; bars show mean and error bars show cross-training-seed standard deviation.
    Our method consistently outperforms all baselines and is the only approach to achieve non-trivial success on bimanual stack-three.}
    \label{fig:main-results}
\end{figure}
\textbf{World-model-aligned BC performance.}
We first compare our GLAM-aligned policy against the baselines discussed in \Cref{sec:exp-setup} on the five manipulation tasks (\Cref{fig:main-results}). GLAM(-O) outperforms all baselines on every task, with the largest absolute gains on the more complex tasks: GLAM-O achieves an average +35\% improvement over the best baseline across three real-world tasks; +44\% on simulated stack-two; and +69\% on bimanual stack-three. Remarkably, on the stack-three task, which typically requires hundreds to a thousand demonstrations to solve~\citep{equivariant,mimicgen}, GLAM(-O) are the only methods that can successfully solve the task ($72.7\%$ vs.\ $\leq 4\%$ for all baselines).
We perform statistical tests on the improvements using the Newcombe-Wilson hybrid-score 95\% confidence intervals~\citep{newcombe} on the difference (GLAM-O $-$ best baseline), which show significant directional improvements across all tasks. 

Both MIP and CLAM(-O) are designed to be single-source only and therefore cannot make use of the auxiliary dataset. As such, these methods tend to overfit to the limited $\mathcal{D}^{\mathrm{tar}}$ and therefore cannot generalise to unseen task configurations.
On the other hand, LAPA(-O) is trained on both sources, but its latent actions are supervised purely by visual reconstruction. 
While prior work ~\citep{lapa} shows that this reconstruction-based approach is sufficient when there is large-scale data, our results demonstrate that IDM reconstruction alone cannot induce reliable transfer between data sources.
In contrast, GLAM's paired ELBOs together with the posterior alignment inject control semantics, yielding a latent space that is source-invariant and control-aware.

\textbf{Object-mask observation improves cross-source transfer.}
\label{sec:ablation}
Overall, we observe that GLAM alone, without object masking, consistently outperforms the baselines. However, as discussed in \Cref{sec:hrwm}, we conjecture that grounding latent actions based on how \textit{objects} should be manipulated can further improve the quality of the learned action space. Here, \Cref{fig:main-results} shows that object-mask input consistently improves task performance across all 5 tasks. In particular, a one-sided Newcombe-Wilson 95\% interval test indicates significant improvements in the knock-down (+25\%) and bimanual stack-three (+16\%) tasks. 
These results corroborate our claim that grounding latent action at the object level can improve transfer across datasets. 

\begin{figure}[t]
    \centering
    \includegraphics[width=0.75\linewidth]{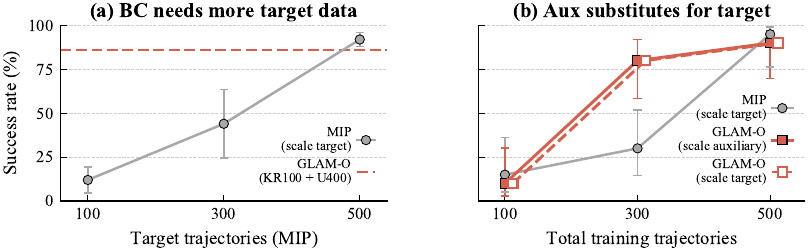}
    \caption{\textbf{Heterogeneous data closes BC's data gap on stack-two.}
    \textbf{(a)} Target-data scaling: MIP needs $5\times$ more target trajectories than GLAM-O to reach the same success rate.
    \textbf{(b)} Auxiliary substitutes for target: for GLAM-O, scaling auxiliary UMI data matches scaling target Kinova data trajectory-for-trajectory (the two curves coincide); MIP scales only with target data.
    Evaluated on 20 fixed unseen initial configurations; error bars are 95\% Wilson intervals.}
    \label{fig:scaling}
\end{figure}

\textbf{Heterogeneous data substitutes for target-robot teleoperation.} 
Our central hypothesis is that using auxiliary data with aligned latent action improves data-efficiency in the target domain. To further investigate this, we perform experiments on the stack-two task with varying amounts of available data to characterise the data-efficiency of the proposed method.
First, we demonstrate that the baseline MIP is able to reliably solve the task when enough target data is available (\Cref{fig:scaling}(a)). Here, by leveraging auxiliary data, GLAM-O is able to match the final performance of the BC baseline with only a small fraction of the target data.
In \Cref{fig:scaling}(b), we present a more fine-grained result on the performance gains offered by unlabelled data, comparing the performance of GLAM-O given a varying mix of target and auxiliary data. Starting from 100 target demonstrations, we compare the performance gains from adding more target data with those from adding auxiliary unlabelled data, up to a total of 500 trajectories.
Note that both cases use the same pretrained GLAM-O model, which is trained on a mixture of target and auxiliary datasets. 
We observe similar scaling behaviour across the two data sources: increasing the amount of auxiliary data yields performance gains comparable to increasing the amount of target data. This supports our claim that the aligned GLAM action space enables auxiliary data to serve as a viable substitute for target data.
Interestingly, \Cref{fig:scaling}(b) also shows that latent policies predicting GLAM latent actions is more data-efficient than vanilla BC. We hypothesize that this is because GLAM has access to more data (500 total episodes) during the pretraining phase which induces a more smooth and informative action space that is more conducive for downstream learning. In \Cref{app:smoothness}, we provide further analysis at the motion-execution level.

\section{Conclusion}
\label{sec:conclusion}

In this paper, we introduce GLAM, a pair of latent-action generative models that leverage heterogeneous demonstrations for supervising downstream imitation learning. GLAM aligns an IDM posterior with an action encoder posterior via paired ELBOs and grounds the action latent in shared forward dynamics, then serves as a frozen latent-action anchor that relabels every transition into source-invariant, control-aware supervision for a BC policy. Trained on a few hundred trajectories without large-scale pretraining, GLAM lets cheap auxiliary data substitute for expensive target teleoperation, alleviating BC's data demand and delivering consistent gains across real-robot and simulated manipulation tasks.
\paragraph{Limitations.}
\label{sec:limitations}
There are several limitations to the current work. First, our auxiliary trajectories share the deployment camera placement and tabletop scene; the framework has not been tested on truly in-the-wild data such as web video~\citep{zeromimic,inNon}, where viewpoint and scene drift would require additional invariance pressure. Second, the framework has not been tested across morphologically distinct end-effectors such as multi-fingered or dexterous hands. A natural extension is to co-train the world model (WM) with human-hand demonstrations~\citep{dexcap,humanplus} as a bridge to multi-fingered embodiments. Third, our auxiliary set shares both the manipulated object and task semantics with the target set; we have not tested auxiliary data sharing only skill primitives with the target task. Since the WM is a general latent-action anchor rather than task-specific, scaling it to many tasks with a shared skill vocabulary, then anchoring single-task policies, is a promising direction. Finally, GLAM grounds the latent in visual object motion, end-effector pose, and proprioception, that suffice for free-space manipulation but cannot capture contact forces or tactile feedback. An interesting extension is to use the WM as a multi-modal bridge: tactile or force signals from instrumented target rollouts~\citep{vtwm,sparsh} flow through the shared latent, lifting vision-only auxiliaries via the unified action space without modality matching on every source.

\clearpage
\acknowledgments{This research was supported by an EPSRC Programme Grant (EP/V000748/1). The authors would like to acknowledge the use of the SCAN computing cluster in carrying out this work. Ingmar Posner holds concurrent appointments as a Professor of Applied AI at the University of Oxford and as an Amazon Scholar. This paper describes work performed at the University of Oxford and is not associated with Amazon.}

\bibliography{lib.bib}  %

\newpage

\appendix
\crefalias{section}{appendix}
\crefname{appendix}{Appendix}{Appendices}
\Crefname{appendix}{Appendix}{Appendices}

\section{Additional Cross-Source Transfer Results}
\label{app:cross-source}
\begin{figure}[h]
    \centering
    \includegraphics[width=\linewidth]{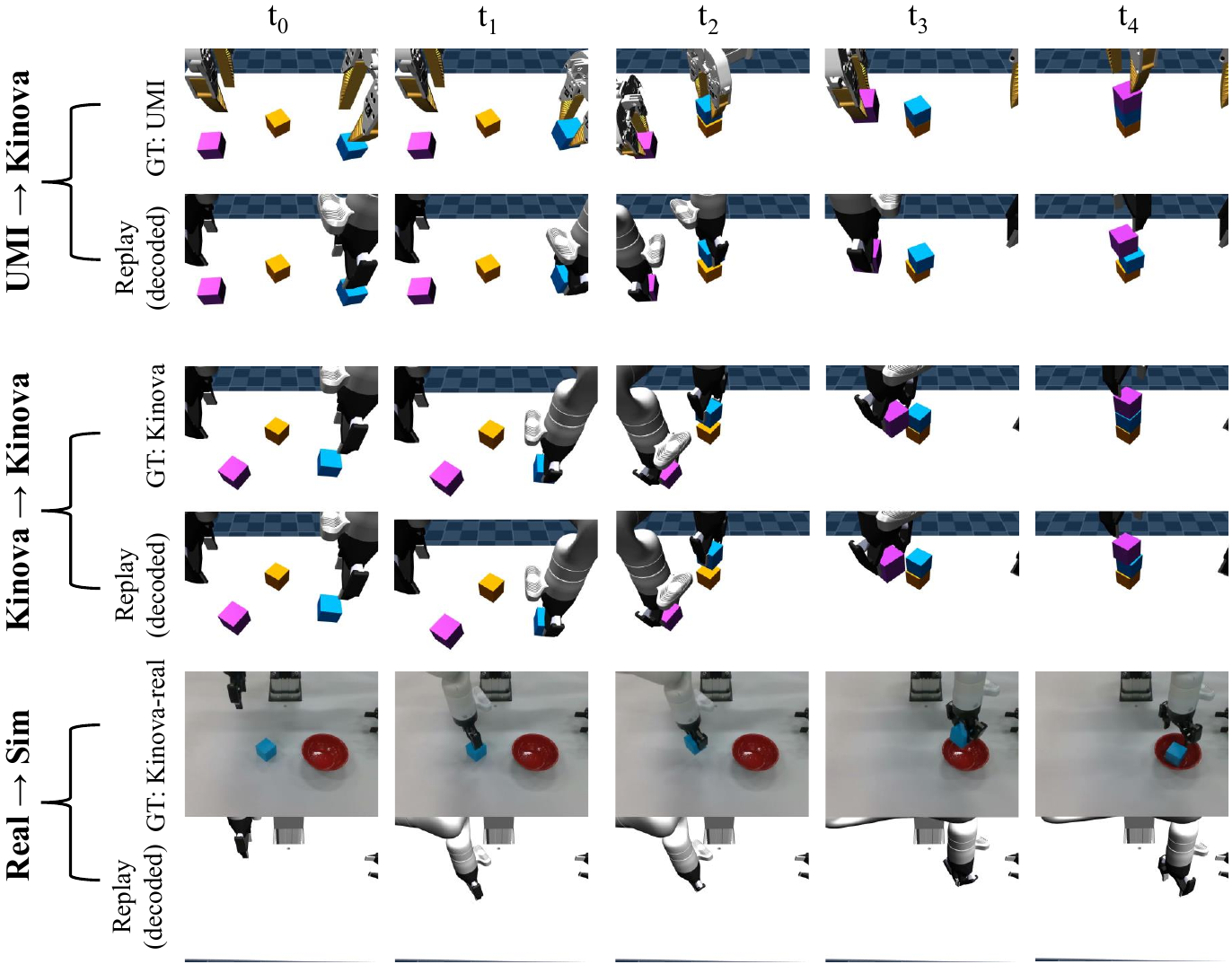} 
    \caption{\textbf{Extended cross-source transfer results.}
    For each row group, an unseen episode (top: GT) is encoded by the IDM into latent actions, decoded by the Kinova-sim action decoder $p_\vtheta(\va_t \mid \vz_t)$, and replayed open-loop on Kinova in simulation (bottom).
\textbf{UMI $\to$ Kinova}: latents inferred from an unseen UMI episode reproduce the manipulation on Kinova, demonstrating cross-embodiment transfer.
\textbf{Kinova $\to$ Kinova}: latents from an unseen Kinova-sim episode replay the original motion, an in-distribution sanity check.
\textbf{Real $\to$ Sim}: latents from an unseen Kinova-real episode transfer to Kinova in simulation, demonstrating sim-real invariance.}
\label{fig:cross-source-transfer-large}
\end{figure}

\Cref{fig:cross-source-transfer-large} extends the qualitative analysis of \Cref{sec:qual-wm} by showing the full episode trajectories. Across UMI, Kinova-sim, and Kinova-real sources, the decoded actions drive the Kinova robot through the entire task open-loop and reach the goal state, indicating that the IDM consistently produces target-executable latent actions for episodes drawn from any source. 

This qualitative experiment of latent quality also guided two design choices in GLAM. First, without the asymmetric alignment in \Cref{eq:align}, the two posteriors collapse toward each other, dragging the action encoder posterior toward the looser, source-mixed IDM; we observed that latents from both the action encoder and the IDM replay unseen Kinova trajectories markedly worse, and UMI latents from the IDM fail to transfer to Kinova. Second, attaching the action-reconstruction term $\log p_\vtheta(\va_t \mid \vz_t)$ to the IDM posterior ${q_\vphi(\vz_t\mid\vx_t,\vx_{t+1})}$, as in CLAM~\citep{clam}, rather than to the action encoder injects a supervision signal that is not unified across sources into the heterogeneous-model IDM posterior (action labels exist only on $\mathcal{D}^{\mathrm{tar}}$), so UMI latents from the IDM fail to transfer to Kinova and the auxiliary data becomes unreliable for downstream BC. Both led to our final design: control semantics enter through the privileged, target-only action encoder, keeping the IDM source-invariant yet decodable into executable actions.

\clearpage

\section{Motion Smoothness Corroborates the Scaling Trends}
\label{app:smoothness}
\vspace{-6pt}
We examine whether the scaling differences in \Cref{fig:scaling} are also visible at the motion level, using two complementary smoothness metrics on the end-effector speed profile $v(t) = \|\dot{\vx}^e_t\|$: a jerk-based score~\citep{LDLJ}, $-\ln|T^5/v_{\text{peak}}^2 \cdot \int_0^T (d^2v/dt^2)^2\,dt|$ (less negative is smoother); and an FFT-based score~\citep{caps,watson}, $S_m = (2/(nf_s))\sum_i M_i f_i$ (lower is smoother). The two metrics come from disjoint mathematical families, time-domain integration vs.\ frequency-domain weighting, so concurrent improvement on both helps avoid measurement artifacts.

\begin{figure}[t]
    \centering
    \includegraphics[width=0.80\linewidth]{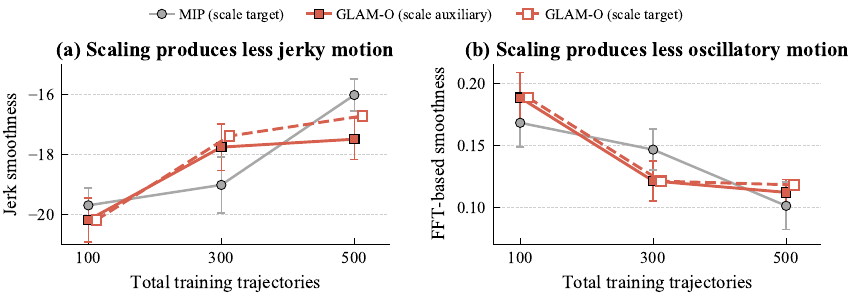}
    \caption{\textbf{End-effector motion smoothness across scaling regimes
    on stack-two}, evaluated on the same 20 unseen initial configurations
    as \Cref{fig:scaling} (b). \textbf{(a)} Jerk-based score~\citep{LDLJ};
    less negative is smoother. \textbf{(b)} FFT-based score~\citep{caps,watson};
    lower is smoother.}
    \label{fig:smoothness}
    \vspace{-6pt}
\end{figure}

\Cref{fig:smoothness} mirrors \Cref{fig:scaling}(b) in three ways.
(i)\textit{ Scaling-curve shape matches success rate.} MIP's smoothness
gain is modest from 100 to 300 trajectories and large from 300 to 500; GLAM-O's is the opposite. Smooth end-effector motion is essential for stable grasping and placement on stack-two, so motion smoothness directly explains the success-rate trends.
(ii)\textit{ GLAM-O is data-efficient at the motion level too.}
The training-scale band over which smoothness sharply improves sits at $300\to500$ for MIP but at $100\to300$ for GLAM-O, matching the data-efficiency gap reported in \Cref{fig:scaling}(b).
(iii)\textit{ Target and auxiliary scaling are kinematically
interchangeable.} The two GLAM-O curves nearly coincide on both metrics, reinforcing the success-rate coincidence in \Cref{fig:scaling}(b) and confirming that auxiliary data can substitute for target data as BC supervision under GLAM.

\begin{figure}[b]
    \centering
    \includegraphics[width=0.9\linewidth]{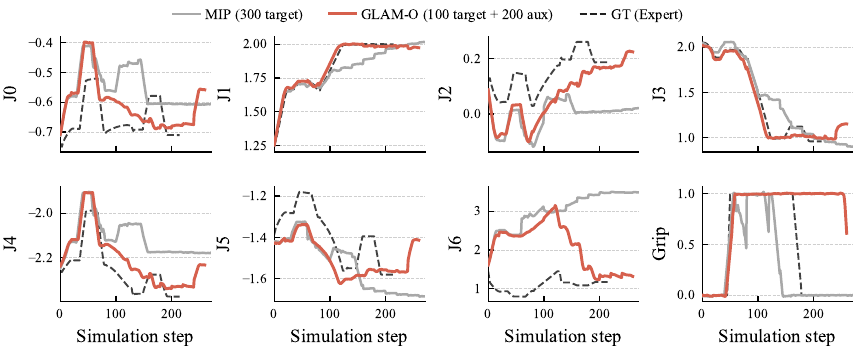}
    \caption{
    Assessing policy generalisation by comparing online joint space rollouts against a validation expert demonstration.
    In this episode,
    MIP fails to stack the two cubes while
    GLAM-O succeeds, despite neither perfectly reproducing the demonstration.
    However, the GLAM-based policy fits the unseen demonstration better and is notably smoother.
    Both policies use 300 total training trajectories, but GLAM-O only uses 100 robot trajectories.}
    \label{fig:joint-trajectories}
\end{figure}
We also inspect the joint-action trajectories to explain the
success-rate gap at 300 trajectories in \Cref{fig:scaling}(b) and \Cref{fig:smoothness}. \Cref{fig:joint-trajectories} shows that GLAM-O's joint trajectories follow the overall trend of the expert better and settle into final configurations consistent with a successful stack; MIP shows abrupt excursions on joint actions (such as J6, which is essential for aligning the end-effector with the cube) and a chattering gripper that never sustains a stable grasp, leading to failure.

\section{Co-training with an Additional Data Source}
\label{app:multi-source}
GLAM's source-invariance claim predicts that adding another data source should help, not hurt. We test this by augmenting the 500-trajectory mix with 100 additional Kinova-sim trajectories (\Cref{tab:third-source}). Co-training improves both pick-and-place ($+3$) and knock-down ($+3$), while lifting holds at $18/20$, already near the per-task ceiling; no task degrades. Because GLAM encodes every source into the same shared latent space, the third source enters the same training pipeline without architectural or weighting changes and contributes additional latent-action labels that enrich the downstream policy's supervision. In principle, the same pipeline accommodates further auxiliary sources across embodiments and domains.
\begin{table}[h]
\centering
\small
\caption{Adding a third source (100 Kinova-sim trajectories) to GLAM training mix maintains or improves all three real-robot tasks without any architectural changes. (KR = Kinova-real, K = Kinova-sim, U = UMI; \emph{e.g.}\ KR100+K100+U400 mixes 100 real-robot + 100 sim-robot + 400 UMI trajectories.)}
\label{tab:third-source}
\begin{tabular}{lccc}
\toprule
\textbf{Task} & \textbf{Two sources} & \textbf{Three sources} & $\Delta$ \\
              & \textit{KR100 + U400} & \textit{KR100 + K100 + U400} & \\
\midrule
Lifting           & \textbf{18/20} & \textbf{18/20} & $0$  \\
Pick \& Place     & 15/20          & \textbf{18/20} & $+3$ \\
Knock-down        & 17/20          & \textbf{20/20} & $+3$ \\
\bottomrule
\end{tabular}
\end{table}

\section{Experimental Setup Details}
\label{app:setup}

\begin{figure}[h]
    \centering
    \includegraphics[width=\linewidth]{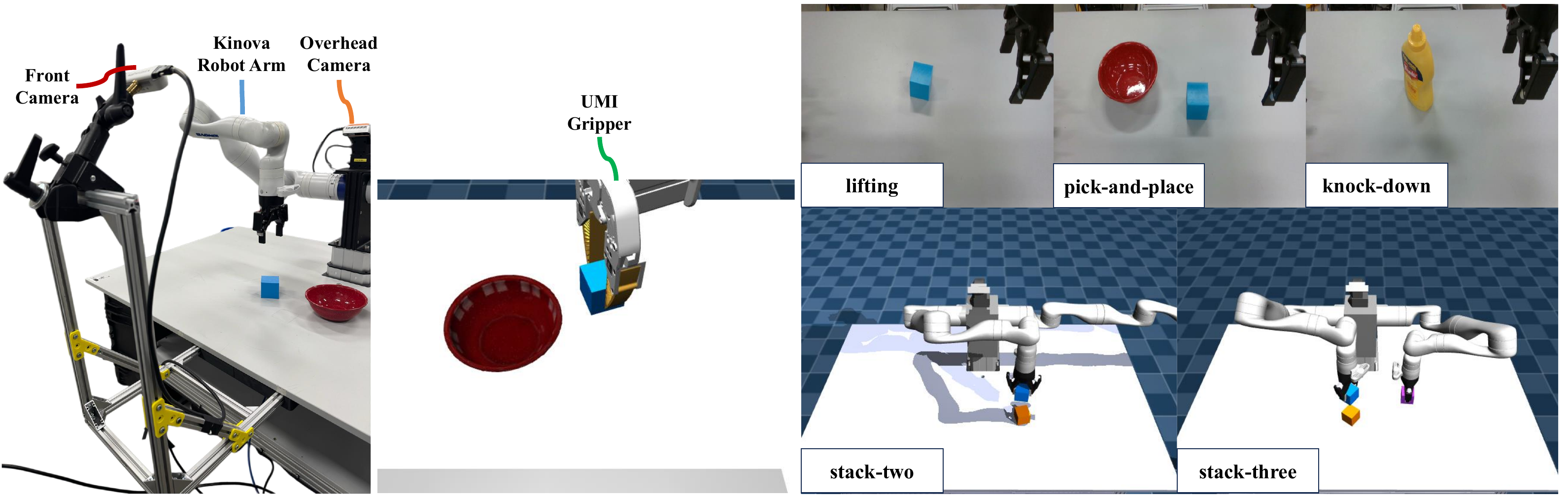} 
    \caption{\textbf{Hardware setup and per-task visualizations.} 
    Left: the Kinova Gen3 with parallel-jaw gripper and two RealSense cameras (overhead, front). 
    Middle: the UMI gripper in MuJoCo for auxiliary demonstrations collection.
    Right: example scenes for each of the five tasks.}
    \label{fig:setup}
\end{figure}

For the real-world tasks, the target platform is a 7-DoF Kinova Gen3 arm with a parallel-jaw gripper, observed by two Intel RealSense cameras (overhead and front). Target demonstrations are collected via teleoperation, while the auxiliary set is generated by scripted policies in simulation, using a UMI gripper~\citep{umi} simulated as a free-floating end-effector (\Cref{fig:setup}: middle). The simulated tasks use MuJoCo (\Cref{fig:setup}: right bottom): the single-arm task reuses the Kinova Gen3 model and the bimanual \emph{three-cube-stacking} task adds a second arm in a shared workspace, both serving as the target robot, while the auxiliary set again uses the UMI gripper. We evaluate each real-world task on 20 unseen object placements and each simulated task on 5 unseen test seeds of 10 rollouts each (50 trials per training seed).

\clearpage

\section{Implementation Details}
\label{app:impl}

This section reports module architectures and hyperparameter values for
the symbols introduced in \Cref{sec:hrwm,sec:policy}.

\paragraph{Module architectures.}
\Cref{tab:arch} summarises all modules. The posteriors $q_\vpsi, q_\vphi, q_\veta$ from \Cref{sec:hrwm} correspond to the visual encoder, IDM, and action encoder, respectively; $p_\vpsi(\vo_t \mid \vx_t)$ is realised by the visual decoder, and $p_\vtheta$ covers the shared forward model, proprio forward, and action decoder. The latent state $\vx_t$ in \Cref{eq:gen-h} is formed by mapping the image part of $\vo_t$ through $q_\vpsi$ to a visual feature $\vy_t \in \sR^{d_o}$ and concatenating with the end-effector pose $\vx_e^t$, giving $\vx_t = (\vy_t, \vx_e^t)$; the image is reconstructed from $\vy_t$ via $p_\vpsi(\vo_t \mid \vx_t)$, and we found that KL regularisation on the latent state was not required in practice. The shared forward model $p_\vtheta(\vx_{t+1} \mid \vx_t, \vz_t)$ uses this same state composition and has a dual head predicting residuals $\Delta\vy_{t+1}$ and $\Delta\vx_e^{t+1}$, which are added to $\vx_t$ to form the next latent state. We replaced the transition-model negative log-likelihood with a deterministic next-state prediction trained by an MSE to simplify the implementation. We additionally sample $\vz_t \sim q_\veta$ during training and pass it through this same forward model, so that both the IDM and action-encoder posteriors drive consistent forward predictions. These two posteriors, $q_\vphi$ and $q_\veta$, share the same Gaussian prior $p(\vz_t) = \gN(\vzero, \mI)$, regularising them into the same bounded region of $\vz$-space. 

The two action decoders, $p_\vtheta(\va_t \mid \vz_t)$ and the policy decoder $\vh^a_\vomega$ (\Cref{eq:policy-loss}), share the same architecture but decode latents from different posteriors: $p_\vtheta(\va_t \mid \vz_t)$ is grounded on action-encoder latents during world-model pretraining, whereas $\vh^a_\vomega$ must decode the IDM latents that relabel the data, so we train it from scratch end-to-end with the policy. We also observed that initialising $\vh^a_\vomega$ from the pretrained $p_\vtheta(\va_t \mid \vz_t)$ yields comparable results, with the action-reconstruction loss converging quickly in either case.

\begin{table}[b]
\centering
\small
\setlength{\tabcolsep}{4pt}
\caption{\textbf{Module architectures of GLAM and the downstream policy.}
``$N$ ResMLP / $H$'' denotes $N$ stacked residual MLP blocks of hidden size 
$H$, each block \textsc{LayerNorm}\,$\to$\,\textsc{Linear}\,$\to$\,\textsc{GELU}\,$\to$\,\textsc{Linear} 
with a residual connection. Conv layers use kernel $4$, stride $2$, padding 
$1$. The MIP policy follows the design of MIP~\citep{mip}.}
\label{tab:arch}
\begin{tabular}{l l l p{5.8cm}}
\toprule
\textbf{Module} & \textbf{Symbol} & \textbf{Backbone} & \textbf{Input $\to$ output} \\
\midrule
\multicolumn{4}{l}{\emph{GLAM world model (\Cref{sec:hrwm})}} \\
\addlinespace[1pt]
Visual encoder   & $q_\vpsi(\vx_t \mid \vo_t)$              & 4 Conv / ReLU             & image $\to \vy_t \in \sR^{d_o}$ \\
Visual decoder   & $p_\vpsi(\vo_t \mid \vx_t)$              & 4 ConvT / ReLU            & $\vy_t \to$ image \\
IDM              & $q_\vphi(\vz_t \mid \vx_t, \vx_{t+1})$   & 5 ResMLP / 256            & $(\vx_t, \vx_{t+1}) \to (\mu, \log\sigma^2) \in \sR^d$ \\
Action encoder   & $q_\veta(\vz_t \mid \vs_t, \va_t)$       & 5 ResMLP / 256            & $(\vs_t, \va_t) \to (\mu, \log\sigma^2) \in \sR^d$ \\
Forward model    & $p_\vtheta(\vx_{t+1} \mid \vx_t, \vz_t)$ & 4 ResMLP / 256, dual head & $(\vx_t, \vz_t) \to \Delta\vy_{t+1},\, \Delta\vx_e^{t+1}$ \\
Proprio forward  & $p_\vtheta(\vs_{t+1} \mid \vs_t, \vz_t)$ & 5 ResMLP / 256            & $(\vs_t, \vz_t) \to \Delta\vs_{t+1}$ \\
Action decoder   & $p_\vtheta(\va_t \mid \vz_t)$            & 4 MLP / 256, ReLU         & $\vz_t \to \va_t$ \\
\midrule
\multicolumn{4}{l}{\emph{Downstream latent-action policy (\Cref{eq:policy-loss})}} \\
\addlinespace[1pt]
Vision encoder   & ---                                      & DINOv2-S/14 + MLP         & Per view DINOv2-S/14 (CLS, $384$-d), jointly trained; 2-view concat $\to$ Linear($768{\to}256$)-LN-GELU-Linear($256{\to}64$)-LN \\
MIP policy       & $\pi_\vomega(\hat\vz_{t:t+H} \mid \vc_t)$ & 10 MLP / 256, LN+Mish    & vision ($64$-d) $\oplus\, \vs_t \to \hat{\vz}_{t:t+H} \in \sR^{H \cdot d}$ \\
Action decoder   & $\vh^a_\vomega$                          & 4 MLP / 256, ReLU         & $\hat\vz_{t:t+H} \to \hat\va_{t:t+H}$ \\
\bottomrule
\end{tabular}
\end{table}

\clearpage
\paragraph{Hyperparameter values.}
\Cref{tab:hparams} lists all hyperparameters. The latent-action dimension $d$ is the only quantity that varies by task. The KL-to-prior weights on the IDM and action encoder posteriors in \Cref{eq:elbo-h,eq:elbo-t} are both set to $10^{-3}$, which prevents posterior collapse and preserves task-relevant information in the latent action.

\begin{table}[h]
\centering
\small
\caption{\textbf{Hyperparameter values for the symbols introduced in
\Cref{sec:hrwm,sec:policy}.}}
\label{tab:hparams}
\begin{tabular}{l l l}
\toprule
\textbf{Symbol} & \textbf{Meaning} & \textbf{Value} \\
\midrule
$d$        & Latent action dim     & $8$; $16$ for stack-three \\
$d_o$      & Visual feature dim    & $128$ \\
$H$        & Policy chunk size     & $20$ \\
\midrule
$\lambda$  & Alignment KL weight (\Cref{eq:total}) & $1$ \\
$-$    & KL-to-prior weight in ELBOs (\Cref{eq:elbo-h,eq:elbo-t}) & $10^{-3}$ \\
\bottomrule
\end{tabular}
\end{table}

\end{document}